\documentclass[twoside,11pt]{article}

\usepackage{jmlr2e}
\usepackage{listings}
\usepackage{xcolor}
\usepackage{authblk}
\usepackage{amsmath}
\usepackage{cleveref}
\usepackage{minted}

\ShortHeadings{DomainLab: A modular Python package for domain generalization in deep learning
}{Xudong Sun, and others}
\firstpageno{1}

\begin{document}

\title{DomainLab: A modular Python package for domain generalization in deep learning
}

\author[1]{Xudong Sun}
\author[2]{Carla Feistner}
\author[3]{Alexej Gossmann}
\author[2]{George Schwarz}
\author[1]{Rao Muhammad Umer} 
\author[2]{Lisa Beer}
\author[4]{Patrick Rockenschaub}
\author[2]{Rahul Babu Shrestha}
\author[1]{Armin Gruber}
\author[8]{Nutan Chen}
\author[1,7,9]{Sayedali Shetab Boushehri}
\author[6]{Florian Buettner}
\author[1,9]{Carsten Marr}
\affil[1]{Institute of AI for Health,
Helmholtz Munich, Munich, Germany}
\affil[2]{Technical University of Munich, Munich, Germany}
\affil[3]{Division of Imaging, Diagnostics, and Software Reliability, CDRH 
U.S Food and Drug Administration, Silver Spring, MD 20993, USA}
\affil[4]{Medical University of Innsbruck, Innsbruck, Austria}
\affil[6]{German Cancer Research Center (DKFZ), Heidelberg, Germany}
%
\affil[7]{Data \& Analytics (D\&A), Roche Pharma Research and Early Development (pRED), Roche Innovation Center Munich, Germany}
\affil[8]{Machine Learning Research Lab, Volkswagen Group, Germany}
\affil[9]{Helmholtz AI, Munich, Germany}


\editor{TBD}

\maketitle

\begin{abstract}
Poor generalization performance caused by distribution shifts in unseen domains often hinders the trustworthy deployment of deep neural networks. Many domain generalization techniques address this problem by adding a domain invariant regularization loss terms during training. However, there is a lack of modular software that allows users to combine the advantages of different methods with minimal effort for reproducibility.
\textit{DomainLab} is a modular Python package for training user specified neural networks with composable regularization loss terms. 
Its decoupled design allows the separation of neural networks from regularization loss construction. Hierarchical combinations of neural networks, different domain generalization methods, and associated hyperparameters, can all be specified together with other experimental setup in a single configuration file. Hierarchical combinations of neural networks, different domain generalization methods, and associated hyperparameters, can all be specified together with other experimental setup in a single configuration file.
In addition, \textit{DomainLab} offers powerful benchmarking functionality to evaluate the generalization performance of neural networks in out-of-distribution data. The package supports running the specified benchmark on an HPC cluster or on a standalone machine. 
The package is well tested with over 95 percent coverage and well documented. From the user perspective, it is closed to modification but open to extension. The package is under the MIT license, and its source code, tutorial and documentation can be found at \url{https://github.com/marrlab/DomainLab}.



\end{abstract}


\section{Introduction}\label{sec:summary}
Deep Learning models suffer from poor generalization when the training and testing distributions are not well aligned~\citep{sun2019high,sun2019variational,gulrajani2020search}. Domain generalization aims at training domain
invariant models that are robust to distribution shifts~\citep{wang2022generalizing}. Implementations of recently published methods in this area are often limited to proof-of-concept code, interspersed with custom code for data access, preprocessing, and evaluation. These custom implementations limit these methods’ applicability to custom datasets, affect their reproducibility, and restrict objective comparison with other state-of-the-art methods.

\textit{DomainBed}~\citep{domainbed2022github}
provided a first codebase for testing domain generalization methods~\citep{gulrajani2020search} but lacks modularity. For instance, each method corresponds to a Python class with all its components hard coded in the initialization method, which results in strong coupling between neural networks and loss functions. From a user perspective, applying its implemented methods to a new use case requires extensive adaptation of its source code across different files, which violates the software design principles of being closed to modification and open to extension~\citep{braude2016software}.


With \textit{DomainLab}, we address the above-mentioned software design issues of \textit{DomainBed} 
by decoupling the components,  which allows combination of different components of domain generalization methods. In addition, we provide a benchmark facility allowing for comprehensive comparisons of domain generalization methods. Our package adheres to  software design patterns and incorporates extensive testing, documentation (function and class level), as well as tutorials. 
\section{Overview of design and features}\label{sec:design_feature}
\subsection{Modularization}\label{sec:modularization}
\textit{DomainLab} was designed to achieve maximum modularity, we introduce the most important modules below.
\paragraph{\emph{Task}}
\emph{Tasks} in \textit{DomainLab} collect datasets from different domains and define the domain generalization scenario via specifying training and test domains. There are
several ways to specify tasks as an input to \textit{DomainLab}: 
\begin{itemize}
    \item \emph{TaskDset} requires users to specify PyTorch Dataset for each domain directly.
    \item \emph{TaskFolder} can be used if the
data is already organized in a folder with different subfolders
containing data from different domains and a further level of
sub-subfolders containing data from different classes.
\item \emph{TaskPathFile} allows the user to specify each domain in a text
file indicating the path and label for each observation so that the user
can choose which portion of the sample to use for training, validation,
and testing. 
\end{itemize}
Additionally, \textit{DomainLab} also provides built-in tasks, e.g.,
Color-MNIST~\citep{sun2021hierarchical}, a subsampled version of PACS~\citep{Li_2017_ICCV}, and of VLCS~\citep{torralba2011unbiased}, to provide a test
utility for algorithms. We provide detailed documentation about \emph{Task} at~ \url{https://marrlab.github.io/DomainLab/build/html/doc_tasks.html}.\\
\paragraph{\emph{Model}s and \emph{Trainer}s} Many domain generalization methods use a Structural Risk Minimization (SRM) like loss $\ell(b(\theta);\xi\sim D_{tr}) + \mu^T \mathcal{R}(b(\theta);\xi \sim D_{tr})$, where $\ell(b(\theta);\xi\sim D_{tr})$ represents the task specific loss (e.g. cross-entropy loss for classification), and $\mathcal{R}(b(\theta);\xi \sim D_{tr})$ represents regularization loss vectors as penalty term to boost domain invariance. Here, $\mu$ represents the multiplier vector to weigh the importance of the two loss terms. We use
$\theta$ to represent neural network weights, and $b(\theta)$ to represent a specific neural network architecture, e.g.~vision Transformer \citep{yuan2021tokens}, ResNet or AlexNet \citep{ heDeepResidualLearning2016,krizhevskyImageNetClassificationDeep2012}, which can map inputs from a data space to a feature space. $\xi\sim \mathcal{D}_{tr}$  represents mini-batch $\xi$ from training domains $\mathcal{D}_{tr}$. There are two ways to construct SRM losses in \textit{Domainlab} using either \emph{Model} or \emph{Trainer} depending on whether auxiliary neural network is needed to construct regularization losses. 

In \textit{DomainLab}, \emph{Models} construct an instance-wise regularization loss $\mathcal{R}(\cdot)$ via auxiliary neural network(s). This  regularization term for \emph{Model}s can be either domain supervised \citep{ilse2020diva, ganin2016domain} 
or domain unsupervised \citep{sun2021hierarchical, carlucci2019domain}.
In contrast, Empirical Risk Minimization (ERM) as a \emph{Model} only has $\ell(b(\theta);\xi\sim \mathcal{D}_{tr})$ but no regularization term.

In \textit{DomainLab},
\emph{Trainer}  is an object that directs data flow to feed into the model for SRM loss calculation, appends further domain invariance losses and updates the model parameters. We formalize this concept in Equation~(\ref{eq:dg}) as $\mathcal{T}_{\theta}$:

\begin{align}
\hat{\theta}=\mathcal{T}_{\theta}\left(\mathbb{E}_{\xi \sim \mathcal{D}_{tr}} [\ell(b(\theta);\xi) + \mu^T  \mathcal{R}(b(\theta);\xi )]; \mathcal{O}_{\theta}, \theta^{(0)}
\right)\label{eq:dg}
\end{align}

Here, $\mathcal{T}_{\theta}$ denotes the \emph{Trainer}, which represents an operator which can add additional regularization loss and bring the model parameter from $\theta^{(0)}$ to $\hat{\theta}$ with optimizer $\mathcal{O}_{\theta}$ by optimizing the penalized loss function $\ell(b(\theta);\xi) + \mu^T \mathcal{R}(b(\theta);\xi )$.\\
Different from \emph{Model}s, \emph{Trainer}s' regularization terms are not constructed upon auxiliary neural networks. For instance,
MatchDG~\citep{mahajan2021domain} uses contrastive learning to distinguish inter-domain similarity from the same class, DIAL~\citep{levi2021domain} creates adversarial samples and augments the training. MLDG~\citep{li2018learning} trains a neural network via splitting training domains into source and target and utilizes Model-Agnostic Meta-Learning (MAML)~\citep{finn2017model}.  FishR~\citep{rame2022fishr} aligns second-order moment information across different domains.



\subsection{Hierarchical combination  across \emph{Trainer}, \emph{Model} and neural network}\label{sec:combo_and_decoration}
The decoupling design of \textit{DomainLab} allows combining \emph{Model} regularization $\mathcal{R}(b(\theta);\xi \sim \mathcal{D}_{tr})$ with \emph{Trainer} $\mathcal{T}_{\theta}$ regularization, as well as decorating \emph{Trainer} with \emph{Trainer}. The decoration and combination feature of \textit{DomainLab} corresponds to extending SRM loss from $\ell(\cdot)+\mu_m R_m(\cdot)$ to $\ell(
\cdot)+\mu_m R_m(\cdot) + \mu_t R_t(\cdot)$ where both $R_m$ (\emph{Model} regularization with dummy index $m$) and $R_t$ (\emph{Trainer} regularization with dummy index $t$) can both be in vector form. We use $\cdot$ to represent $b(\theta); \xi\sim D_{tr}$ for brevity of expression. Such combination and decoration can be done recursively to have the form $\ell(
\cdot)+\sum_k \mu_{m_k} R_{m_k}(\cdot)+\sum_j \mu_{t_j} R_{t_j}(\cdot)$ by combining the regularization loss from \emph{Trainer}s (indexed by $t_j$) and \emph{Model}s (indexed by $m_k$). See section \ref{sec:trainer_model_separation} for a use case. Additionally, we provide an example of \emph{Model} decorating \emph{Model} in \url{https://github.com/marrlab/DomainLab/blob/v0.2.4/examples/api/jigen_dann_transformer.py} where the neural network is a vision transformer~\citep{yuan2021tokens}. 
\subsection{Benchmark}
\textit{DomainLab} supports benchmarking the performance of methods against a custom domain generalization scenario, by taking systematic and stochastic variation into consideration. Within a \textit{YAML} configuration file, the user can specify common experimental settings (e.g., a common neural network backbone $b(\theta)$) and individual hyperparameter ranges for each method to be benchmarked. Then the whole benchmark can be executed by a single command, where hyperparameters and random seeds get sampled, with automatic aggregation and visualization of results, which offers more performance distribution information compared to presenting results in a table. Powered by \textit{Snakemake} \citep{molderSustainableDataAnalysis2021}, \textit{DomainLab} supports job submission to an HPC cluster or a standalone machine. 

\section{Use case}
\subsection{Combination and decoration between \emph{Trainer} and \emph{Model} }\label{sec:trainer_model_separation}
To decorate the \emph{Trainer} MLDG~\citep{li2018learning} with DIAL~\citep{levi2021domain} (see feature explanation in \Cref{sec:combo_and_decoration}) to train the \emph{Model} DIVA~\citep{ilse2020diva}, one could use the following command:
\begin{small}
\begin{minted}{shell}
domainlab --config=./examples/conf/vlcs_diva_mldg_dial.yaml 
\end{minted}
\end{small}
where the configuration file below with detailed comments can be downloaded at \begin{small}
\url{https://github.com/marrlab/DomainLab/blob/master/examples/conf/vlcs_diva_mldg_dial.yaml}\end{small}
\begin{center}
\begin{small}
\begin{minted}[linenos]{yaml}
te_d: caltech                       # name of test domain from VLCS dataset
tpath: examples/tasks/task_vlcs.py  # python file path to specify the task 
bs: 2                               # batch size
model: dann_diva                    # combine Model DANN with DIVA
epos: 1                             # number of epochs
trainer: mldg_dial                  # decorate Trainer MLDG with DIAL
gamma_y: 700000.0                   # hyperparameter of DIVA
gamma_d: 100000.0                   # hyperparameter of DIVA
npath: examples/nets/resnet.py      # neural network for class classification
npath_dom: examples/nets/restnet.py # neural network for domain classification
\end{minted}
\end{small}
\end{center}

\subsection{Benchmarking domain generalization algorithms upon a custom dataset}\label{sec:ex_benchmark}
By simply editing a \textit{YAML} configuration file, users can specify the domain generalization task,  define the hierarchical combination of \emph{Trainer}, \emph{Model} and neural net along with their respective hyperparameter ranges. Hyperparameters can be sampled either randomly or with fixed grids, according to specified distributions. In addition, methods can also share common hyperparameters. In this case, a pool of shared hyperparameter sets is sampled, from which each method takes its hyperparameter samples, ensuring consistency in the hyperparameter selection.
The following command runs the benchmark on a \textit{Slurm} HPC cluster.
\begin{small}
\begin{minted}{shell}
bash run_benchmark_slurm.sh examples/benchmark/demo_shared_hyper_grid.yaml
bash \end{minted}
\end{small}

\begin{figure}[htbp!]
\centering
\includegraphics[width=0.8\textwidth]{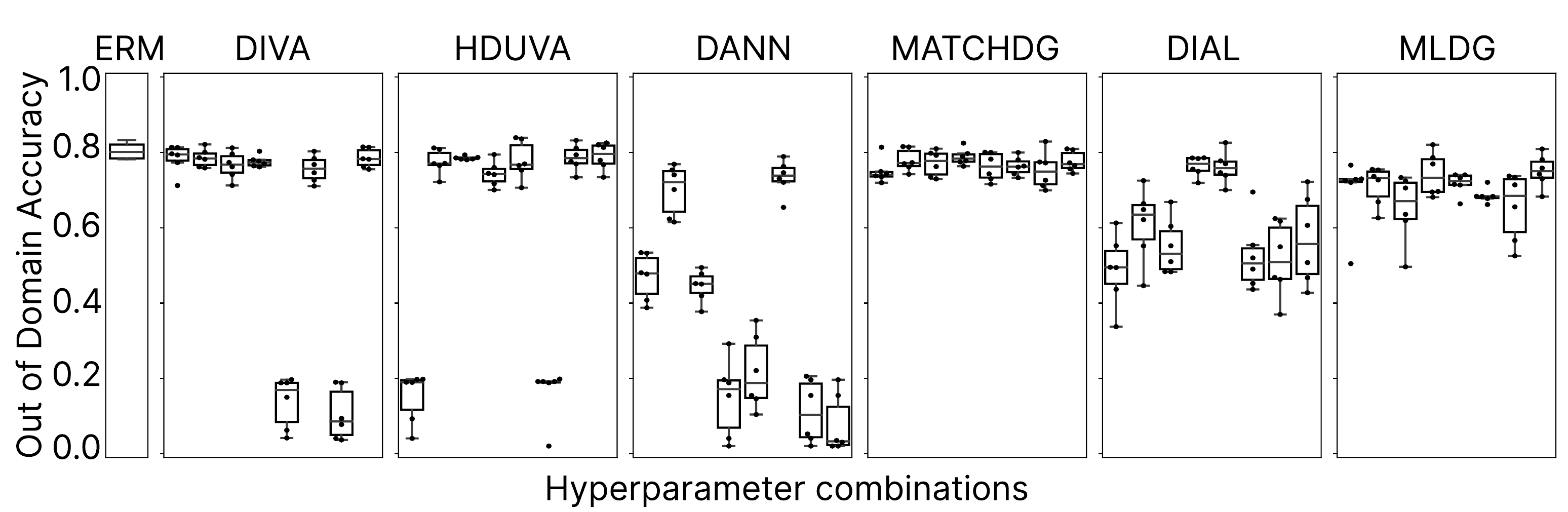}
\caption{
\textit{DomainLab} benchmarking out-of-domain generalization performance using accuracy by testing on the \textit{sketch} domain on the PACS dataset~\citep{Li_2017_ICCV} based on an adapted version of ResNet50 from ~\citep{gulrajani2020search} in conjunction with different domain generalization methods, where ERM (empirical risk minimization) is the baseline. Each panel corresponds to a domain generalization method (\textit{Models} and \textit{Trainers} from \Cref{sec:modularization}). From the figure, ERM remains a competitive baseline for domain generalization, as found in \citep{gulrajani2020search}.
Hyperparameter configuration for domain generalization algorithm affects the generalization performance or even result in failed training. 
}
\label{fig:pacs}
\end{figure}

The example benchmark configuration file can be found in \url{https://github.com/marrlab/DomainLab/blob/master/examples/benchmark/demo_shared_hyper_grid.yaml}
After all the scheduled benchmark jobs are finished, \textit{DomainLab} will visualize the benchmark results via various plots. For instance, Figure~\ref{fig:pacs} shows the benchmark results on the \textit{PACS}~\citep{Li_2017_ICCV} dataset, where we use the \textit{Sketch} domain as the test domain. The generated visualizations enable the comparison of different algorithms and the assessment of the effects of hyperparameter changes and random seed variations. Further explanations can be found in the figure caption, as well as the detailed documentation in \url{https://marrlab.github.io/DomainLab/build/html/doc_benchmark.html}.

\section{Conclusion}
\textit{DomainLab} is a thoroughly tested and well-documented software platform for training domain invariant neural networks. It follows software design patterns with a focus on maximum decoupling among its components. The package allows hierarchical combinations of different methods in domain generalization and provides a powerful benchmark function which can be executed on either a \textit{Slurm} cluster or a standalone machine. 
\newpage
\appendix
\section{Author Contributions}\label{contributions}
XS designed and implemented the architecture and framework of the package with software design patterns, implemented each domain generalization method, designed and implemented the model trainer combination mechanism, documentation and website generation script, and other major package components and facilities, initiated and made significant contributions to other aspects of the package development. AGo participated in regular discussion on some design aspects of the early phase software components. AGo co-authored some code development with XS, maintained software facility such as documentation generation, CI, testing through the use of DomainLab in independent research
projects, contributed with patches, fixes and code review, and helped writing the manuscript. GS designed and implemented the possibility to benchmark different algorithms using a \textit{Snakemake}
pipeline, implemented the random sampling of hyperparameters and contributed minor enhancements. 
CF added grid search of hyperparameters, implemented shared hyperparameter sampling and the chart generation for the graphical evaluation of the benchmark results. CF further implemented and fixed numerous small features like sanity checks for \emph{Task} and many more. CF improved the manuscript. CM initiated the project
with XS, contributed to the code style enhancement, improved the paper description
of Domainlab, and supervised the project. 
RU assisted XS in implementing examples with vision transformer and implementation of evaluation metrics, unit testing, benchmarking testing on the cluster and reviewed the manuscript.
LB contributed by writing documentation for \emph{Model}s and \emph{Trainer}s and assisted XS in adding unit tests for user API coding. 
PR contributed to the package testing, code quality, proofread and improved the paper. 
RS added a feature to specify command line
arguments with YAML files. 
AGr made a use case of the package with a real-world medical datasets and implemented the \emph{Task} for the blood cell classification problem. 
NC introduced precommit into the package, improved code formatting and proofread the paper. SSB improved several files' code formatting under the help of XS. FB contributed to the code documentation on \emph{Model} HDUVA and improved the manuscript. CM initiated the project
with XS, contributed to the code style enhancement and paper description
of Domainlab, and supervised the project. 


\acks{We thank Mariia Sidulova for building open source projects \citep{sidulova2023deep} on top of \textit{DomainLab}. 
We thank Xinyue Zhang (who we did not manage to reach out) for offer code printing and saving of the confusion matrix.
CM has received funding from the
European Research Council (ERC) under the European Union's Horizon 2020
research and innovation program (Grant agreement No.~866411).}

\newpage
\bibliography{bib_domainlab_jmlr_mloss}

\end{document}